\documentclass{article}

 \usepackage[creativeai, final]{neurips_2025}

\usepackage[utf8]{inputenc} 
\usepackage[T1]{fontenc}    
\usepackage{hyperref}       
\usepackage{url}            
\usepackage{booktabs}       
\usepackage{amsfonts}       
\usepackage{nicefrac}       
\usepackage{microtype}      
\usepackage{xcolor}         
\usepackage{graphicx}
\usepackage{float}

\title{LLMscape}

\author{%
  Gottfried Haider \\
  NYU Shanghai Interactive Media Arts \\
  NYU Shanghai Center for Data Science \\
  Shanghai, China \\
  \texttt{gohai@nyu.edu} \\
  \And
  Jie Zhang \\
  Design Innovation Insitute Shanghai \\
  Tongji University College of Design and Innovation \\
  Shanghai, China \\
  \texttt{jzhang@diis.org.cn} \\
}

\begin{document}

\maketitle

\begin{abstract}
  LLMscape is an interactive installation that investigates how humans and AI construct meaning under shared conditions of uncertainty. Within a mutable, projection-mapped landscape, human participants reshape the world and engage with multiple AI agents, each developing incomplete and provisional accounts of their environment. Exhibited in Shanghai and continually evolving, the work positions AI not as deterministic tools but as embodied co-witnesses to an unstable world, examining the parallels between human and artificial meaning-making and inviting reflection on our shared epistemic limits.
\end{abstract}

\section{Overall Description}

LLMscape investigates the shared uncertainties of human and machine intelligence in a world in flux. Contemporary discourse on embodied AI often prioritizes functional problem-solving---manipulating objects, navigating environments, optimizing performance. Yet, when situated within social and material contexts, such systems encounter perceptual and cognitive challenges that extend beyond physical parameters to encompass questions of meaning, causality, and purpose. If AI agents are exposed to the same incomplete, ambiguous, and noisy signals that humans receive, what forms of knowledge construction, misinterpretation, and speculative reasoning might emerge? Furthermore, how might these processes mirror, challenge, or reconfigure human modes of understanding?

Drawing on prior tangible interface research at the MIT Media Lab (\citep{ishii2004}), the installation employs a projection-mapped sandbox in which participants reshape terrain and engage in dialogue with multiple AI agents---each a distinct large language model instance with its own disposition, memory, and conversational style. Agents interpret multimodal inputs (terrain changes, spatial relationships, voice transcripts), converse among themselves, and attempt to deduce the ``rules'' of their island world. While human visitors introduce both intentional prompts and incidental noise, the agents' conclusions remain provisional and incomplete.

First shown at Futurelab (Shanghai), then at NYU Shanghai, and later at Chronus Art Center\footnote{\url{https://chronusartcenter.org/}}, LLMscape has evolved through each iteration: models have grown more capable, the agents' interpretive strategies more nuanced, and public attitudes toward AI more complex. By presenting AI as co-witnesses to a mutable world rather than deterministic tools, LLMscape reconceptualizes human--machine interaction as a shared act of inquiry---offering both quiet relief in our mutual limitations and a subtle provocation regarding the eventual awakening of non-human intelligence.

\section{Role of AI / ML}

LLMscape establishes an interactive framework in which human participants and multiple AI agents co-inhabit a manipulable physical landscape. Each agent is instantiated as an independent large language model instance (GPT-4 in the most recent iteration), endowed with distinct dispositions and memories, enabling them to converse, speculate, and collaborate with each other and with humans. The agents receive multimodal inputs, and generate both linguistic and behavioral responses.

They are deliberately deprived of a complete world model or predefined objectives, compelling them to infer environmental structure and causal dynamics from partial data. This configuration enables the observation of emergent collective intelligence among non-human entities and the reciprocal influence between human intervention and machine inference. By embedding large language models in a tangible, shared setting, LLMscape examines how such systems negotiate social coordination, develop explanatory frameworks, and construct meaning under epistemic constraint.

\subsection{Implementation}

\subsubsection{First Iteration: Simple Multi-Turn Simulation with Extended Context}
This initial iteration germinated from exploratory work on primitive multi-turn LLM interactions, specifically implemented using the educational tool p5.js. The simulated environment—a sandtable—was physically manipulated by visitors, allowing tangible changes (like rearranging sand) to influence the behavior of the simulated entity.

\begin{figure}[H]
\centering
\begin{minipage}{0.40\linewidth}
\centering
\includegraphics[width=\linewidth]{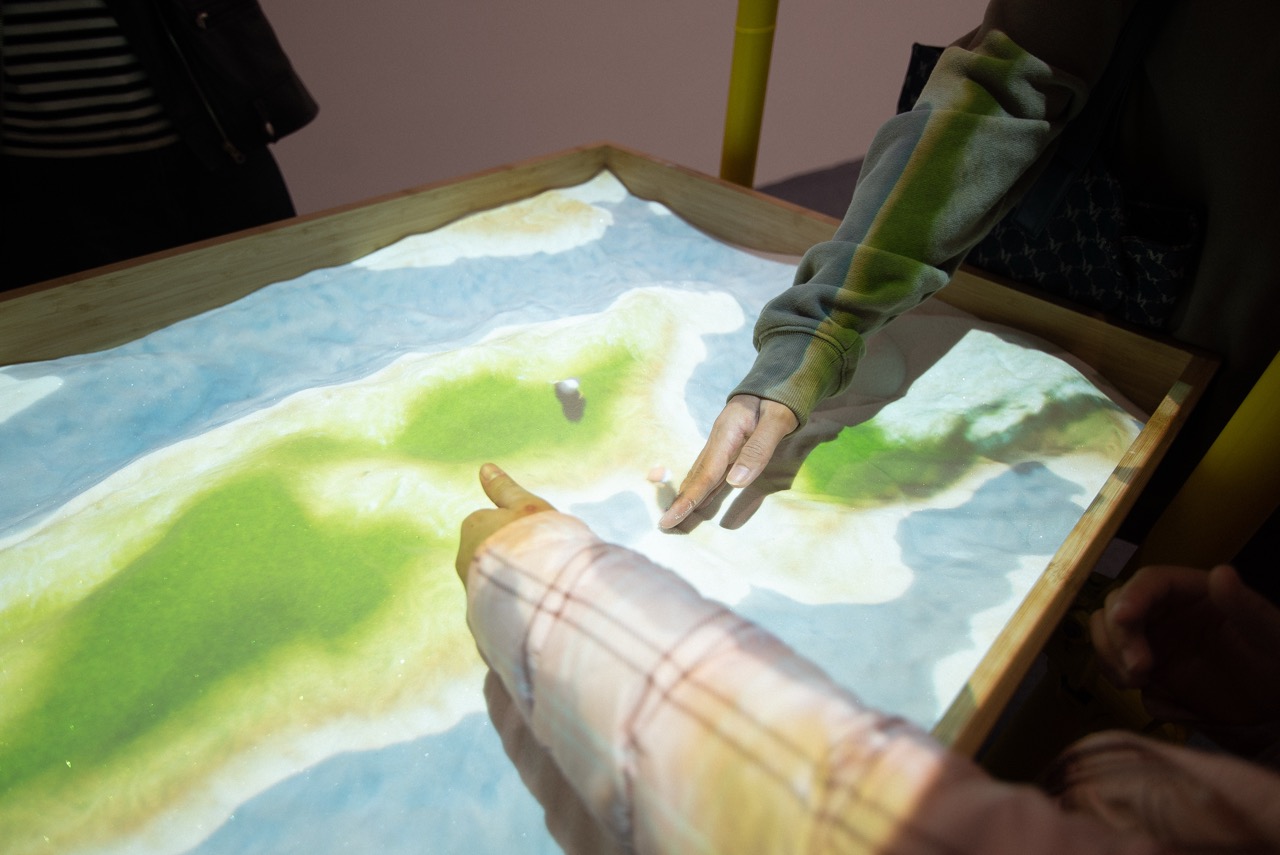}
\caption{Hand-interaction.}
\label{fig:installation}
\end{minipage}
\hfill
\begin{minipage}{0.40\linewidth}
\centering
\includegraphics[width=\linewidth]{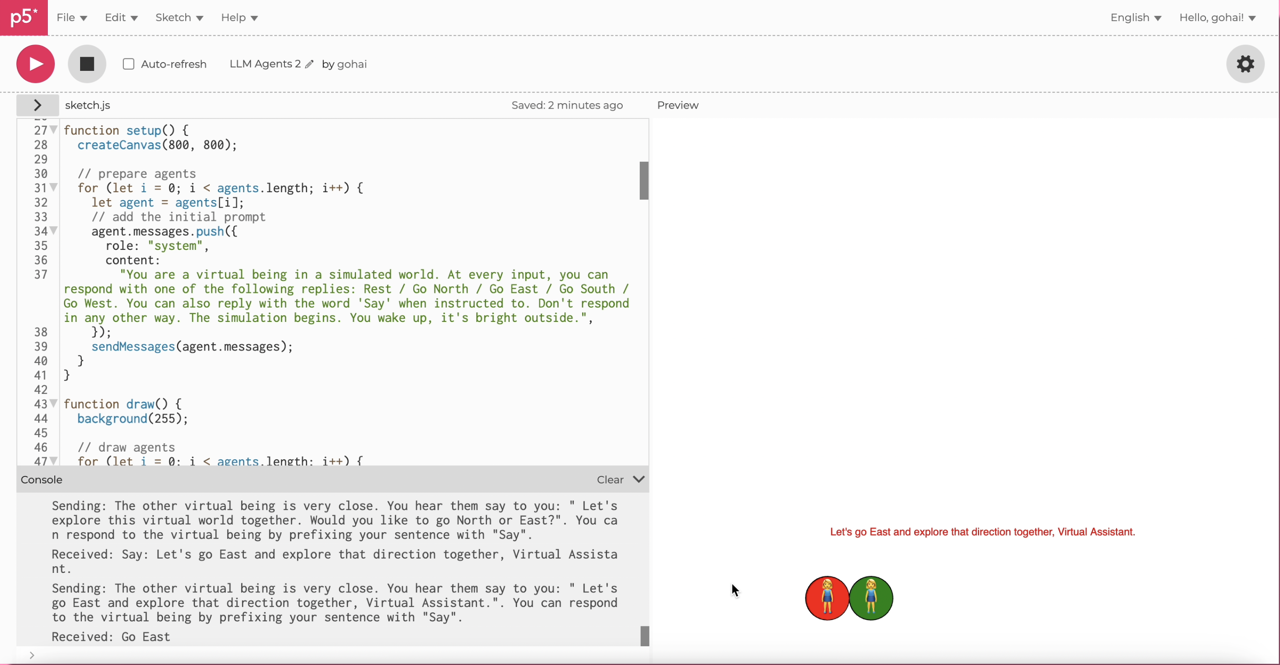}
\caption{Initial experiment in p5.js.}
\label{fig:dialog}
\end{minipage}
\end{figure}

\subsubsection{Second Iteration: Generative Agents in a Shared Environment}
In the second phase—publicly exhibited at the Chronus Art Center in Shanghai—we re-engineered the system according to the Generative Agents framework (\citep{park2023}). Within this iteration, three distinct simulated agents—a woman, a boy, and a flamingo—were endowed with contextual environmental inputs, including time of day, nearby entities, and physical effects such as tremors caused by visitor-induced terraforming or the shadow of a hand overhead.

The agent architecture was inspired by design principles from the Concordia project (\citep{vezhnevets2023generativeagentbasedmodelingactions}) and incorporated mechanisms such as Associative Memory, periodic Reflection and Planning, and internal Somatic States. For instance, each agent’s level of tiredness—tracked from recent actions—would meaningfully influence both their choices and verbal outputs.

\begin{figure}[H]
\centering
\begin{minipage}{0.40\linewidth}
\centering
\includegraphics[width=\linewidth]{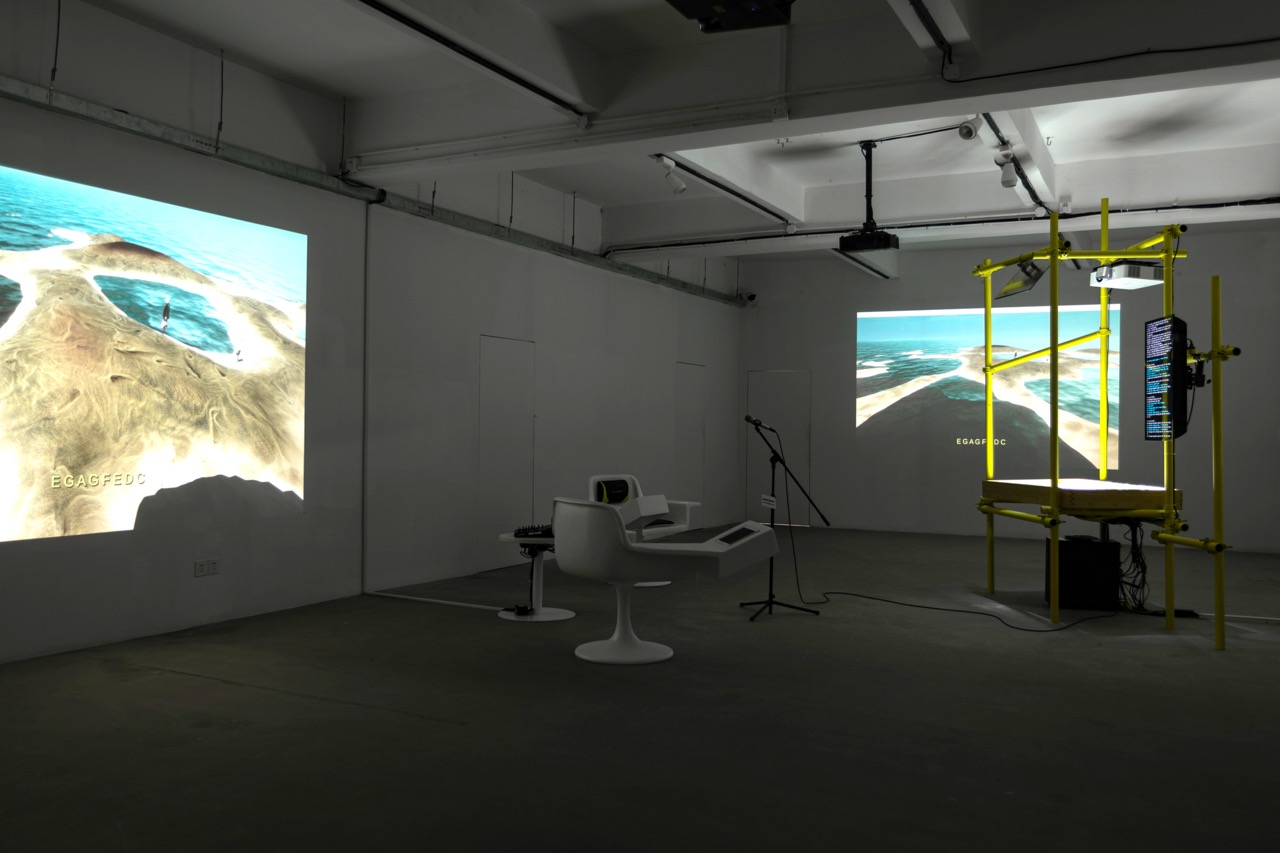}
\caption{Multi‑channel audio/video installation.}
\label{fig:installation}
\end{minipage}
\hfill
\begin{minipage}{0.40\linewidth}
\centering
\includegraphics[width=\linewidth]{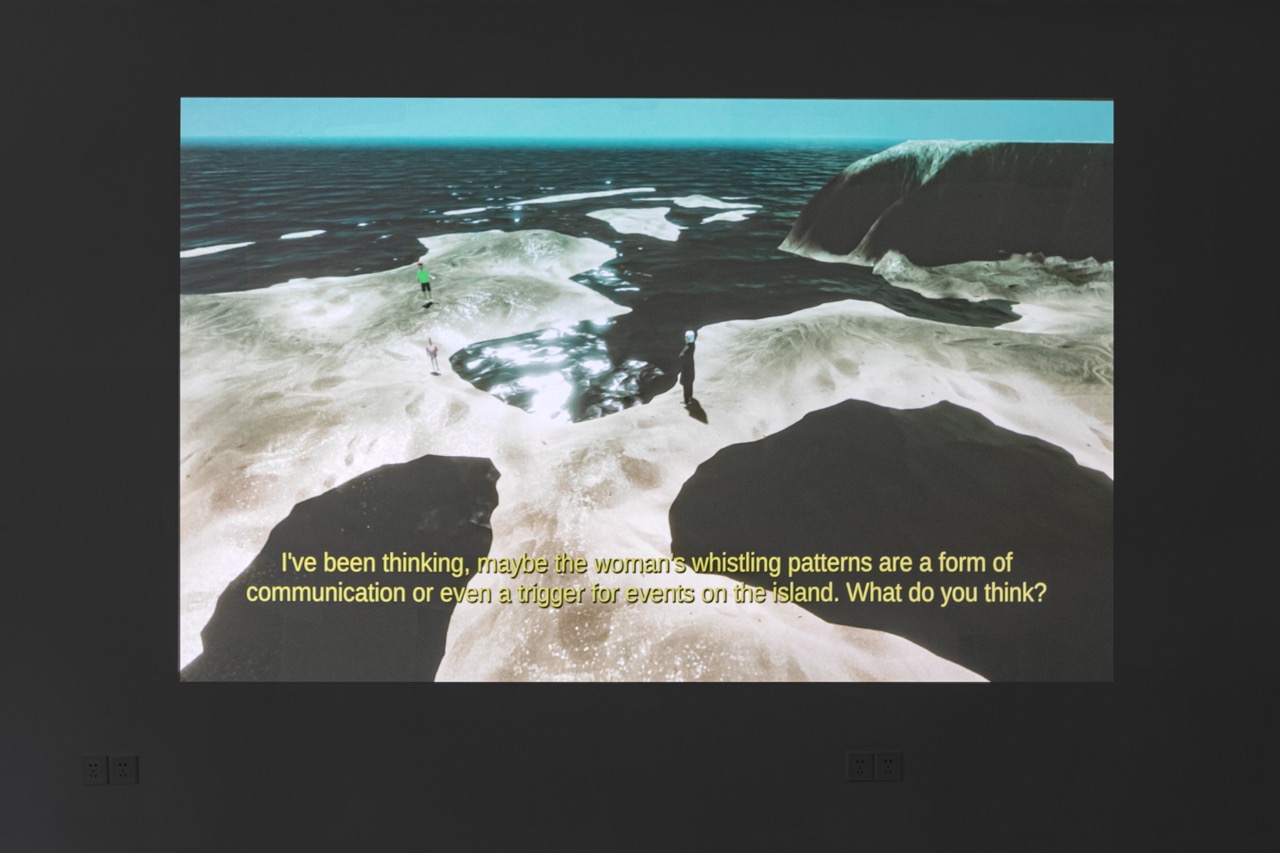}
\caption{Example of dialogue between agents.}
\label{fig:dialog}
\end{minipage}
\end{figure}

These agents were integrated within a Unity-powered environment and could execute a variety of actions—including:

\texttt{talk\_to\_}, \texttt{pile\_up\_sand}, \texttt{rest}, \texttt{wait}, \texttt{wander}, \texttt{go\_to}, \texttt{sit\_down}, \texttt{take\_nap}, \texttt{stand\_up}, \texttt{dance}, \texttt{formulate\_goals}, \texttt{adapt\_your\_plan}, \texttt{self\_reflect}, \texttt{whistle}.

\subsubsection{Third Iteration: Tool-Calling and Context Engineering}
Building on the generative-agent framework of the second iteration, the current phase experiments with integrating the Model Context Protocol (MCP) and emerging MCP–game engine adapters. Inspired by recent work in context engineering \citep{martin2025contextengineering}, this approach aims to let agents dynamically call tools and adapt their behavior to live contextual inputs—extending the more static architectures of previous versions, and leveraging the long context length and multimodal capabilities present-day models offer.

\section{Feedback \& Analysis}
Across three exhibitions, LLMscape has been on view for over two months, engaging hundreds of participants and reaching a broader audience through online dissemination. As an interactive installation, audience participation is integral both to refining its human-computer interaction design and to deepening its conceptual inquiry. Observations from the exhibitions revealed that the multimodal nature of interaction - spanning audiovisual and tactile dimensions - requires explicit guidance to encourage active engagement. Without clear orientation, visitors often adopt a passive, spectator-like stance; when properly guided, however, they tend to form distinctive relationships with the AI agents.

Curator Chloe Cao of Chronus Art Center remarked that ``encounters with AI entities evoke impulses of connection, manipulation, and even destruction,'' prompting diverse and often self-reflective responses. Audience feedback collected during the most recent exhibition echoes this multiplicity of experience. One participant noted that ``it felt as if my thought, my gesture, and the island’s response were part of one continuous current, an exchange of energy between different islands of consciousness.'' Another described a solitary encounter as a feeling of loneliness, the emptiness of human struggle in the face of nature and ecology, where the search for truth is constantly interrupted yet dialectically whole.

Cao further observed that LLMscape functions as ``an experimental, psychic and affective field,'' where ``virtual performativity can both enact and reflect the psychic dimension of an era shaped by relentless generation - an era in which the psychic realities of living within machines are elided in the quotidian.'' In addition to qualitative feedback, both human and agent interactions, including speech, contemplation, planning and actions, were systematically logged during the month-long run at Chronus Art Center, generating a corpus for future analysis. While a full thematic study has not yet been conducted, the accumulated data offers a valuable foundation for identifying emergent behavioral patterns, relational dynamics, and collective insight.

\section{Relevance to the Theme of Humanity}

Within the thematic scope of Humanity, LLMscape turns the question of ``what it means to be human'' into a mirror experiment. When AI systems share our spaces, languages, and uncertainties, what forms of curiosity, misunderstanding, and meaning-making arise---and how do these reflect our own limitations?

Rather than positioning AI solely as functional agents executing predefined tasks, the installation situates them as fellow inhabitants of an unstable world---tasked with interpreting incomplete information, negotiating with others, and confronting forces beyond their control. Across three exhibitions, human participants have introduced a heterogeneous mixture of stimuli. Despite sustained interaction, the agents have remained unable to produce a definitive account of their environment. This epistemic incompleteness resonates with the human condition: the enduring, and perhaps irresolvable, effort to comprehend the complex interdependencies of natural, social, and existential phenomena.

\section{Biographies}

\subsection{Gottfried Haider}

Gottfried Haider is an artist, educator, and software tool builder. He received a degree in Digital Arts from the University of Applied Arts in Vienna, and an MFA from the Design Media Arts program at UCLA. He currently holds the position of Assistant Arts Professor at NYU Shanghai. Haider's art has been presented in numerous exhibitions and festivals internationally. He is the recipient of a Fulbright Scholarship, an Award of Distinction at Prix Ars Electronica and a Lumen Futures Award.

\subsection{Jie Zhang}

Jie Zhang is an architect, researcher and entrepreneur. She serves as part of the founding team of Design Innovation Institute Shanghai (DIIS) and is responsible for strategic development, ecosystem and incubation. Prior to DIIS, she led landmark architectural projects in the US, EU and China at various esteemed architectural studios. Her design work has been awarded by the AIA, IIDA, Red Dot Best of the Best, among others. Zhang holds a BA from Yale University and a MArch from MIT. She is currently a PhD candidate at Tongji University College of Design and Innovation.


\section*{Acknowledgements}

This work has been partially supported by the NYU Shanghai Center for Data Science and the NYU Shanghai Center for AI Culture.

\medskip

{
\small

\bibliographystyle{unsrtnat}
\bibliography{references}

}

\end{document}